\documentclass[11pt,a4paper]{article}
\usepackage{hyperref}
\usepackage{acl2019}
\usepackage{times}
\usepackage{latexsym}
\usepackage{amsmath,amssymb,amsfonts}
\usepackage{algorithmic}
\usepackage{graphicx}
\usepackage{textcomp}
\usepackage{xcolor}
\usepackage{adjustbox}
\usepackage{multirow}
\usepackage{makecell}
\usepackage{url}
\usepackage[LFE,T1,LAE,T2A]{fontenc}
\usepackage[utf8]{inputenc}
\usepackage[main=english,farsi]{babel}

\aclfinalcopy 
\def\aclpaperid{} 

\title{\textbf{NSURL-2019 Task 7: Named Entity Recognition (NER) in Farsi}}

\author{
\textbf{Nasrin Taghizadeh, Zeinab Borhanifard, Melika GolestaniPour, Heshaam Faili} \\
School of Electrical and Computer Engineering, College of Engineering, \\
University of Tehran, Tehran, Iran \\
\texttt{nsr.taghizadeh@ut.ac.ir} 
  }

\date{}
\hypersetup{draft}
\begin{document}
\maketitle
\begin{abstract}
NSURL-2019 Task 7 focuses on Named Entity Recognition (NER) in Farsi. This task was chosen to compare different approaches to find phrases that specify Named Entities in Farsi texts, and to establish a standard testbed for future researches on this task in Farsi. This paper describes the process of making training and test data, a list of participating teams (6 teams), and evaluation results of their systems. The best system obtained 85.4\% of F$_1$ score based on phrase-level evaluation on seven classes of NEs including person, organization, location, date, time, money and percent.
\end{abstract}

\section{Introduction}

Named Entity Recognition (NER) is defined as the task of identifying relevant nouns such as persons, products, and genes which are mentioned in a text. NER is an important task as it is usually employed as a primary step in other tasks such as event detection from news, customer support for on-line shopping, knowledge graph construction, and biological analysis \cite{bokharaeian2017snpphena}.

NER is a famous and well-studied task in English \cite{yadav2018survey} and some other languages like Arabic \cite{shaalan2014survey,helwe2019arabic,taghizadeh2018cross} and German \cite{riedl2018named}. However, this task is not highly examined in Farsi because there is no standard benchmark for it. Although there is some Farsi NER corpus such as PEYMA \cite{Shahshahani2018PEYMAAT}, ArmanPersoNER \cite{poostchi2016personer}, A’laam\cite{Shekofteh}, and Persian-NER\footnote{\url{https://github.com/Text-Mining/Persian-NER}}; none of them is known as standard data set to the research community. Moreover, the type of named entities and annotation guidelines are different in each corpus. Because of the diversity of annotation types and data sets which were used for training and test, results of current researches on Farsi NER cannot be directly compared.

The goal of this competition was to bring Farsi NER researchers together. We introduce a large scale corpus containing about 900K tokens as the training data for this task. To evaluate the participating teams, a test set was prepared which contains 150K tokens. The training and test set follow the same annotation schema. These data sets are publicly available for further researches\footnote{\url{https://github.com/nasrin-taghizadeh/NSURL-Persian-NER}}. The domain of all data is the news sentences because they are the most entity-rich.

Participants were allowed to use any public data and resources such as Farsi Wikipedia\footnote{\url{https://fa.wikipedia.org/wiki/}} and Farsi Knowledge Graph\footnote{\url{http://farsbase.net/search/html/index.html}} \cite{sajadi2018farsbase} in addition to the official training data of the shared task in the process of making their system. In this case, they must thoroughly describe those resources and the way they used them. 

To the best of our knowledge, this is the first shared task in Farsi. Since Farsi belongs to the group of low-resource languages \cite{taghizadeh2016automatic,fadaei2019using}, the availability of annotated corpora and resources will be very useful for future investigation in this language.

\section{Farsi NER}
So far, some researchers have been conducted on Farsi NER. Poostchi et al. \cite{poostchi2018bilstm} presented a BiLSTM-CRF model, which is a recurrent neural network obtained by a combination of a long short-term memory (LSTM) and a conditional random field (CRF). They presented a public data set for Farsi NER, called ArmanPersoNER, which includes six types of NEs: person, organization, location, facility, product, and event. Their model showed 77.45\% of F$_1$ on ArmanPersoNER.

Shahshahani et al. \cite{Shahshahani2018PEYMAAT} presented a hybrid system consisting of a rule-based and a statistical system. The rule-based system composed of a large list of NEs in Farsi in addition to the regular expressions for detecting them. The statistical system is a CRF model trained by the PEYMA corpus. Their system reached 84\% of F$_1$ for seven classes of person, organization, location, date, time, money, and percent, based on 5-fold validation on the training data.

Hossinnejad et al. \cite{Shekofteh} presented a corpus named A’laam consisting of 13 classes of named entities. They split this corpus into two parts of 90\% and 10\% for the training and test, respectively, and trained a CRF model using the training part. They obtained 92.9\% and 78.5\% of precision and recall, respectively.

\textbf{Zafarian et al. \cite{zafarian2015semi} proposed a semi-supervised method for Farsi NER. They used an un-labeled bilingual data in addition to a small labeled data to train their system. They presented a bootstrap method that iteratively trains a CRF model using the labeled data as well as those un-labeled data that the current model predicts them with high confidence. Their data contains three classes of person, organization, and location. They reached 67.5\% of F$_1$.}

Current researches on Farsi NER use different data for the training and test. Most of these data are not public or annotated with diverse annotation schema. The evaluation methods of them are not similar and so their results cannot be directly compared.

\section{The Task}
Participating systems have to predict NE tags for a set of tokenized documents. We defined two subtasks: 
\begin{itemize}
\item
3-classes including person, organization, and location;
\item
7-classes including date, time, money, and percent in addition to the three above classes.
\end{itemize}
 
NEs that belong to four classes of date, time, money, and percent sometimes can be recognized using the rule-based or hybrid methods \cite{ahmadi2015hybrid,riaz2010rule}; while NEs belong to the classes of person, organization, or location are often recognized based on the gazetteer lists and they are more subject to ambiguity. Therefore, we have separated these two subtasks and participants could submit different systems for them.

\subsection{Baseline Method}

CoNLL 2003 defined the baseline of NER task a system that only selects complete unambiguous named entities that appear in the training data. We adapted this baseline as follows: 
\begin{itemize}
\item
In case of overlap between two candidate named entities, the longer one is selected. For example, consider three NEs of the training data: 1) ``\FR{ایران}/Iran'' which is a \textit{location}, 2) ``\FR{مجلس شورای اسلامی}/Islamic Consultative Assembly'' which is an \textit{organization}, and 3) ``\FR{مجلس شورای اسلامی ایران}/Islamic Consultative Assembly of Iran'' which is an \textit{organization} as well. To extract named entities from the phrase ``\FR{مجلس شورای اسلامی ایران}/Islamic Consultative Assembly of Iran'', the baseline system selects the whole phrase as an \textit{organization} instead of separately tagging ``\FR{مجلس شورای اسلامی}/Islamic Consultative Assembly'' as an \textit{organization} and ``\FR{ایران}/Iran'' as a \textit{location}. 

\item
When two NEs are next to each other and have the same tag, they are merged. For example, there are two NEs in the training data: ``\FR{۲۲ بهمن}/22th of Bahman'' and ``\FR{۱۳۵۷}/1357'' which are \textit{date}. In the test phase, the baseline system visits phrase ``\FR{۲۲ بهمن ۱۳۵۷}/22th of Bahman 1357'', and separately selects these two phrases as \textit{date}. Then, they are merged to be one mention. Our analysis showed that this heuristic is often true. However, in a few cases, it may be wrong. For example, consider the following sentence:
\begin{flushright}
\FR{گزینه اول جانشینی آمانو کورنل فروتا دیپلمات رومانیایی است.}
\end{flushright}
``The first option for Amano's successor is Cornel Fruta, a Romanian diplomat."

There are two adjacent mentions with the same \textit{person} type: ``\FR{آمانو}/Amano'' and ``\FR{کورنل فروتا}/Cornel Feruta'', and merging them into one NE is not correct.

\end{itemize}
These examples reveal some challenges of Farsi NER. One challenge is that a unique named entity may appear in the text with different names. For example, ``\FR{مجلس شورای اسلامی ایران}/Islamic Consultative Assembly of Iran'', ``\FR{مجلس شورای اسلامی}/Islamic Consultative Assembly'' and ``\FR{مجلس}/assembly'' are different names of the same entity. While ``\FR{مجلس}/assembly'' is a common noun, it names an \textit{organization}. It means that gazetteers are not sufficient for detecting boundaries of entity mentions. 

Another challenge is that two or more entity mentions may be adjacent in the sentence, in the sense that there is no word between them. They may have different or similar types. In case of similar types, it may be possible or not to merge them into a unique mention. For example adjacent entity mentions of \textit{date}, mostly can be merged, such as ``\FR{بیست و هفتم مهر ماه سال 1398}".

\section{Data Set Creation}
\label{sec:data}

We presented a training data set which has two parts: the first part is PEYMA corpus \cite{Shahshahani2018PEYMAAT} containing 300K tokens; the second part has 600K tokens. The same annotation schema was used for annotating two parts. This annotation schema was prepared based on two standard guidelines: 1) MUC\footnote{\url{https://www-nlpir.nist.gov/related_projects/muc/proceedings/ne_task.html}} and 2) CoNLL\footnote{\url{https://www.clips.uantwerpen.be/conll2003/ner/}}; then it was adapted for Farsi linguistic structures \cite{Shahshahani2018PEYMAAT}. In these data sets there are seven classes of named entities: \textit{person}, \textit{organization}, \textit{location}, \textit{money}, \textit{date}, \textit{time}, and \textit{percent}. 

Steps of creating data set include news collection, pre-processing, and named entity tagging. The test data has two parts: in-domain and out-of-domain. The former was sampled from the same news websites in the same period of time that the training data were collected. The latter was selected from different news websites at different times. Specifically, documents of the training data mostly were sampled from few Farsi news websites between 2016 and 2017; while out-of-domain documents were sampled from many Farsi news websites from different countries of the world mainly in 2019.
Therefore, in-domain documents are more similar in word distribution to the training data than the out-of-domain documents. 

Pre-processing on news documents was performed using Persianp toolkit \cite{DBLP:conf/cicling/MohseniGF16}, which includes tokenization, sentence split, and normalization. Two annotators performed the annotation task, and the agreement between them is 95\% which shows the quality of the annotations. The data format is similar to the CoNLL 2003, in which each line contains one word and empty lines represent sentence boundaries. Annotation format is IOB that encodes the beginning and inside of the entity mentions and type of them.

\begin{table}[tbp]
\caption{Data Statistics}
\begin{adjustbox}{width=\columnwidth}
\begin{tabular}{|l|c|c|c|c|}
\hline 
 & Lang & \#Doc & \#Sent & \#Tokens\\ 
\hline 
Training Data & Fa & 1,456 & 27,130 & 885,296 \\ 
\hline 
Test Data & Fa & 431 & 4,154 & 144,526 \\ 
\hline 
\hline
ArmanPersoNER & Fa & - & 7,682 & 250,015 \\
\hline
CoNLL-2003 & En & 1,393 & 22,137 & 301,418 \\
\hline
CoNLL-2003 & Gr & 909 & 18,973 & 310,318 \\
\hline
\end{tabular}
\label{tb:data-statistics}
\end{adjustbox}
\end{table}

\begin{table}
\caption{Statistics of Test data}
\begin{adjustbox}{width=\columnwidth}
\begin{tabular}{|c|c|c|c|}
\hline 
Test Data & \#Doc & \#Sent & \#Tokens \\ 
\hline 
In-domain & 196 & 1,571 & 68,063  (47\%)\\ 
\hline 
Out-of-domain & 235 & 2,583 & 76,463 (53\%) \\ 
\hline 
\end{tabular} 
\end{adjustbox}
\label{tb:in-out-domain}
\end{table}

\begin{table}[tbp]
\caption{Number of total phrases tagged per class}
\begin{adjustbox}{width=\columnwidth}
\begin{tabular}{|c|c|c|c|c|c|c|c|c|}
\hline 
Data & PER & ORG & LOC & MON & DAT & TIM & PCT & Total \\ 
\hline 
Training & 12,495 & 14,205 & 15,403 & 1,294 & 4,467 & 571 & 997 &  49,432\\ 
\hline 
Test & 2,738 & 3,160 & 4,081 & 357 & 1,147 & 165 & 156 & 11,804 \\ 
\hline 
\end{tabular} 
\end{adjustbox}
\label{tb:per-class}
\end{table}

\begin{table}[tbp]
\caption{Number of unique phrases tagged per class}
\begin{adjustbox}{width=\columnwidth}
\begin{tabular}{|c|c|c|c|c|c|c|c|c|}
\hline 
Data & PER & ORG & LOC & MON & DAT & TIM & PCT & Total \\ 
\hline 
Training & 5,228 & 4,547 & 2,738 & 1,008 & 1,910 & 338 &  453 & 16,020 \\ 
\hline 
Test & 1,470 & 1,326 & 1,015 & 288 & 628 & 114 &  97 & 4,917 \\ 
\hline 
\end{tabular} 
\end{adjustbox}
\label{tb:per-class-unique}
\end{table}

\subsection{Data Statistics}
Table \ref{tb:data-statistics} represents general statistics of our Farsi data sets including the number of articles, sentences, and tokens in comparison with English and German data sets of the CoNLL 2003. The comparison reveals that the Farsi training data is a large scale data set that can be used for further researches on Farsi NER. Table \ref{tb:in-out-domain} shows details of the in-domain and out-of-domain parts of the test data. The two parts have a nearly equal number of tokens. Tables \ref{tb:per-class} and \ref{tb:per-class-unique} represent the total number of phrases and the number of unique phrases tagged for each class of named entities in the training and test data. Considering the size of each corpus, the test set is more dense in terms of the entity tags.

\begin{table*}[tbp]
\caption{Description of Participating Systems}
\begin{center}
\begin{adjustbox}{width=\textwidth}
\begin{tabular}{|l|c|c|c|}
\hline 
Team & Model & Word Embeddings & Features \\ 
\hline 
MorphoBERT & BERT + BiLSTM & \makecell[c]{BERT for token representation\\word2vec for word clustering} & \makecell[c]{cluster number of words,\\morphology} \\ 
\hline 
Beheshti-NER-1 & Transformer-CRF & BERT & - \\ 
\hline 
Team-3 & CRF & - & - \\ 
\hline 
ICTRC-NLPGroup & CRF & - & n-gram, lemma, linguistics rules \\ 
\hline 
UT-NLP-IR & CRF & - & \makecell[c]{POS, NP-chunk, word n-gram,\\ char n-gram, stem, lemma} \\ 
\hline 
SpeechTrans & SVM & - & \makecell[c]{word unigram, char 5-grams,\\ POS, stem, normalized surface} \\ 
\hline 
Baseline & heuristic & - & - \\
\hline
\end{tabular} 
\end{adjustbox}
\label{tb:descrb}
\end{center}
\end{table*}

\section{Participating Systems and Results}
Six teams have participated in both subtasks. Most of them opted for use of CRF models and deep learning methods specifically Bi-LSTM. Because these two models deal with sequence tagging problems. Word embeddings, n-grams, and POS tags were used as features by the systems. Morphological and orthographic features of Farsi phrases were used by some of the participants. Table~\ref{tb:descrb} briefly shows the models and features used by the participants.


\subsection{Evaluation Metrics}
There are different methods for the evaluation of NER systems. Two main methods are phrase-level and word-level evaluation. In the phrase-level evaluation, a phrase is counted as true-positive for class c, if both boundaries of the phrase and its predicted tag are correct. In contrast, in word-level evaluation, each word is considered separately. Therefore, the phrase-level evaluation is more tough than the word-level evaluation. 

We used the evaluation script of conlleval\footnote{\url{https://github.com/sighsmile/conlleval}}. This script computes three measures including precision, recall and F$_1$ based on the standard definition. Evaluation of the 3-classes subtask has been performed based on the macro-averaging method. Accordingly, precision and recall are obtained by averaging of the precision and recall of the three classes of \textit{person}, \textit{organization}, and \textit{location}. 

Evaluation of the 7-classes subtask has been conducted using the micro-averaging method due to class imbalance problem, in the sense that frequencies of NE phrases belonging to four classes of \textit{date}, \textit{time}, \textit{money}, and \textit{percent} are very fewer than the three classes of \textit{person}, \textit{organization} and \textit{location}, according to the Tables \ref{tb:per-class} and \ref{tb:per-class-unique}. So, in this case, micro-averaging better evaluates the quality of systems.

\subsection{Result}

Participating teams mainly used sequence tagging methods including CRF and Bi-LSTM networks. The feature sets used by them include lexical, morphological, and structural features. Tables \ref{tb:eval-3-test1} and \ref{tb:eval-7-test1} show the evaluation results of 3-classes and 7-classes subtasks, respectively. Generally, results of the word-level evaluation are higher than the phrase-level evaluation. Moreover, the results of the evaluation by the in-domain data are higher than the out-of-domain data in terms of F$_1$ score. 
All teams outperformed the baseline and ranking of the teams is the same based on all kinds of evaluations. 


The best F$_1$ scores are 85.9\% and 88.5\% based on the phrase-level and word-level evaluation, respectively, which are obtained by MorphoBERT system \cite{Task7:NERFarsi-mohseni}. The second best system, Beheshti-NER-1 \cite{Task7:NERFarsi-taher}, got near F$_1$ scores: 84.0\% and 87.9\% based on the phrase-level and word-level evaluation, respectively. These two systems used BERT model \cite{devlin2018bert} for training high accurate representation of Farsi tokens. BERT is a deep bi-directional language model that presented state-of-the-art results in a wide variety of NLP tasks. Both systems used the BERT to process a huge amount of un-labeled Farsi texts to obtain pre-trained word embeddings which then was fine-tuned for the NER task 

MorphoBERT used a morphological analyzer as a prior step before the BERT network. Farsi is rather rich-morphology and analyzing tokens to find their parts reveals the grammatical and semantic information. So, instead of embedding tokens of sentences into the network, MorphoBERT firstly decomposes tokens into constituents and then fed these constituents into the BERT network. Then, the representation of the sentence which was obtained from the BERT is given to a Bi-LSTM network. Additionally, a vector representing word cluster features is given to the Bi-LSTM. Finally, the Softmax layer produces a probability distribution over all classes \cite{Task7:NERFarsi-mohseni}.
 
Beheshti-NER-1 system utilizes a CRF model on top of the BERT network. The motivation of using CRF is that an encoder like BERT tries to maximize the likelihood by selecting best-hidden representations, and CRF tries to maximize the likelihood by selecting best output tags \cite{Task7:NERFarsi-taher}.

\begin{table*}[htbp]
\caption{Evaluation of systems for subtask 3-classes}
\begin{center}
\begin{adjustbox}{width=\textwidth} 
\begin{tabular}{|c|l|c c c|c c c|c c c|c c c|c c c|c c c|}
\hline
\multicolumn{2}{|c}{} & \multicolumn{18}{|c|}{\textbf{Test Data}} \\
\cline{3-20} 
\multicolumn{2}{|c}{} & \multicolumn{9}{|c}{\textbf{Phrase-level evaluation}} & \multicolumn{9}{|c|}{\textbf{Word-level evaluation}} \\
\cline{3-20}
\multicolumn{2}{|c}{Team} & \multicolumn{3}{|c}{\textbf{\textit{In-domain}}}& \multicolumn{3}{|c}{\textbf{\textit{Out-of-domain}}}& \multicolumn{3}{|c}{\textbf{\textit{Total}}} & \multicolumn{3}{|c}{\textbf{\textit{In-domain}}}& \multicolumn{3}{|c}{\textbf{\textit{Out-of-domain}}} & \multicolumn{3}{|c|}{\textbf{\textit{Total}}} \\
\cline{3-20} 
\multicolumn{2}{|c|}{} & P & R & F$_1$ & P & R & F$_1$ & P & R & F$_1$ & P & R & F$_1$ & P & R & F$_1$ & P & R & F$_1$ \\
\hline
1 & MorphoBERT & 88.7 & 85.5 & 87.1 & 86.3 & 83.8 & 85.0 & 87.3 & 84.5 & 85.9 & 92.5 & 86.7 &  89.5 & 91.5 & 84.0 & 87.6 & 92.1 & 85.2 & 88.5 \\
\hline
2 & Beheshti-NER-1 & 85.3 & 84.4 & 84.8 & 84.4 & 82.6 & 83.5 & 84.8 & 83.3 & 84.0 & 90.5 & 87.2 & 88.8 & 89.7 & 85.0 & 87.3 & 90.1 & 85.8 & 87.9 \\
\hline
3 & Team-3 & 87.4 & 77.2 & 82.0 & 87.4 & 73.4 & 79.8 & 87.4 & 75.0 & 80.7 & 89.2 & 79.5 & 84.1 & 89.5 & 74.7 & 81.4 & 89.3 & 76.9 & 82.7 \\
\hline
4 & ICTRC-NLPGroup & 87.5 & 76.0 & 81.3 & 86.2 & 69.6 & 77.0 & 86.8 & 72.3 & 78.9 &  90.1 & 78.2 & 83.7 & 88.7 & 70.2 & 78.4 & 89.4 & 73.5 & 80.7 \\
\hline
5 & UT-NLP-IR & 75.3 & 68.9 & 72.0 & 72.3 & 60.7 & 66.0 & 73.6 & 64.1 & 68.5 & 87.3 & 71.9 & 78.9 & 86.4 & 61.1 & 71.6 & 86.9 & 65.7 & 74.8 \\
\hline
6 & SpeechTrans & 41.5 & 39.5 & 40.5 & 43.1 & 38.7  & 40.8 & 42.4  & 39.0 & 40.6 & 66.8 & 38.3 & 48.7 & 66.2 & 35.2 & 46.0 & 66.6 & 36.4 & 47.0 \\
\hline
7 & Baseline & 32.2 & 45.8 & 37.8 & 32.8 & 39.1 & 35.7 & 32.5 & 41.9 & 36.6 & 46.2 & 42.6 & 44.3 & 45.2 & 35.1 & 39.5 & 45.9 & 38.4 & 41.8 \\
\hline
\end{tabular}
\end{adjustbox}
\label{tb:eval-3-test1}
\end{center}
\end{table*}

\begin{table*}[htbp]
\caption{Evaluation of systems for subtask 7-classes}
\begin{center}
\begin{adjustbox}{width=\textwidth}
\begin{tabular}{|c|l|c c c|c c c|c c c|c c c|c c c|c c c|}
\hline
\multicolumn{2}{|c}{} & \multicolumn{18}{|c|}{\textbf{Test Data}} \\
\cline{3-20} 
\multicolumn{2}{|c}{} & \multicolumn{9}{|c}{\textbf{Phrase-level evaluation}} & \multicolumn{9}{|c|}{\textbf{Word-level evaluation}} \\
\cline{3-20}
\multicolumn{2}{|c}{Team} & \multicolumn{3}{|c}{\textbf{\textit{In-domain}}}& \multicolumn{3}{|c}{\textbf{\textit{Out-of-domain}}}& \multicolumn{3}{|c}{\textbf{\textit{Total}}} & \multicolumn{3}{|c}{\textbf{\textit{In-domain}}}& \multicolumn{3}{|c}{\textbf{\textit{Out-of-domain}}} & \multicolumn{3}{|c|}{\textbf{\textit{Total}}} \\
\cline{3-20} 
\multicolumn{2}{|c|}{} & P & R & F$_1$ & P & R & F$_1$ & P & R & F$_1$ & P & R & F$_1$ & P & R & F$_1$ & P & R & F$_1$ \\
\hline
1 & MorphoBERT & 88.4 & 84.8 & 86.6 & 86.0 & 83.1 & 84.5 & 87.0 & 83.8 & 85.4 & 94.0 & 89.1 & 91.5 & 91.8 & 85.7 & 88.6 & 92.8 & 87.1  & 89.9 \\
\hline
2 & Beheshti-NER-1 & 84.8 & 83.6 & 84.2 & 83.9 & 82.0 & 83.0 & 84.3 & 82.7 & 83.5 & 91.4 & 87.3 & 89.3 & 89.7 & 85.7 & 87.7 & 90.4 & 86.5 & 88.4 \\
\hline
3 & Team-3 & 87.4 & 77.3 & 82.0 & 87.3 & 72.8 & 79.4 & 87.3 & 74.7 & 80.5 & 91.3 & 84.1 & 87.5 & 90.9 & 77.9 & 83.9 & 91.1  & 80.7 & 85.5 \\
\hline
4 & ICTRC-NLPGroup & 87.0 & 76.1 & 81.2 & 86.2 & 70.2 & 77.4 & 86.5 & 72.7 & 79.0 & 89.2 & 83.1 & 86.1 & 89.8 & 76.5 & 82.6 & 89.7 & 79.4 & 84.2 \\
\hline
5 & UT-NLP-IR & 77.3 & 70.2 & 73.6 & 74.1 & 61.9 & 67.5 & 75.5 & 65.4 & 70.1 & 92.7 & 79.3 & 85.4 & 91.1 & 68.4 & 78.1 & 91.9 & 73.1 & 81.4 \\
\hline
6 & SpeechTrans & 38.0 & 34.5 & 36.2 & 38.9 & 33.6 & 36.0 & 38.5 & 34.0 & 36.1 & 76.1 & 32.9 & 45.9 & 74.9 & 30.3 & 43.2 & 75.7 & 31.5 & 44.5 \\
\hline
7 & Baseline & 32.8 & 45.7 & 38.2 & 32.0 & 38.1 & 34.8 & 32.4 & 41.3 & 36.3 & 50.6 & 47.8 &49.2  & 42.6 & 35.1 & 38.5 & 46.5 & 40.9 & 43.5 \\
\hline
\end{tabular}
\end{adjustbox}
\label{tb:eval-7-test1}
\end{center}
\end{table*}

To better understand the details of the scores, we presented the F$_1$ scores of each 7 classes based on the phrase-level evaluation in Table \ref{tb:f1-per-class}. Generally, the most F$_1$ scores were obtained by \textit{percent} and \textit{money} classes. Because there are specific keywords representing them and so there are high-precision patterns that specify entity mentions of these classes. Specifically, \textit{percent} often comes with the keywords like ``\FR{درصد}/percent''; while \textit{money} appears with words and phrases denoting money like
``\FR{دلار}/Dollar'', ``\FR{ریال}/Rial'', or ``\FR{یورو}/Euro''. On the other hand, the least F$_1$ scores were obtained by the \textit{time} class. Perhaps because the number of phrases in the training data having \textit{time} tag is very few in comparison to the other classes.


\begin{table}
\caption{Details of phrase-level evaluation for subtask 7-classes (values are F$_1$ score)}
\begin{center}
\begin{adjustbox}{width= \columnwidth}
\begin{tabular}{|c|l|c c c c c c c |c|}
\hline
\multicolumn{2}{|c}{\multirow{2}{*}{Team}} & \multicolumn{7}{|c|}{Named Entity Classes} & \multirow{2}{*}{F$_1$} \\
\cline{3-9} 
\multicolumn{2}{|c|}{}  & PER & ORG & LOC & DAT & TIM & MON & PCT &  \\
\hline
1 & MorphoBERT & 90.4 & 80.3 & 87.1 & 78.9  & 71.0 & 93.6 & 96.8 & 85.4 \\
\hline
2 & Beheshti-NER-1 &  81.8 & 80.8  & 88.0 & 77.8 & 75.8 & 85.1 & 91.6 & 83.5 \\
\hline
3 & Team-3 & 79.9 & 77.2 & 83.9  & 74.7 & 64.3 & 92.1 & 97.4 & 80.5 \\
\hline
4 & \footnotesize{ICTRC-NLPGroup} & 76.2 & 75.93 & 82.8 & 76.0 & 67.1 & 91.3 & 93.6 & 79.0\\
\hline
5 & UT-NLP-IR & 63.4 & 58.8 & 78.2 & 76.1 & 69.1 & 84.5 & 93.5 & 70.1 \\
\hline
6 & SpeechTrans & 24.3 & 23.5 & 63.1 & 12.0 & 4.1 & 0.3 & 0.7 & 36.1 \\
\hline
7 & Baseline & 23.5 & 38.1 & 44.2 & 41.6 & 30.3 & 13.7 & 36.6 & 36.3 \\
\hline
\end{tabular}
\end{adjustbox}
\label{tb:f1-per-class}
\end{center}
\end{table}

\section{Conclusion}
We have described the NSURL-2019 task 7: NER in Farsi. Six systems have processed the Farsi NE data. The best performance was obtained by the MorphoBERT system that is 85.4\% of F$_1$ score based on the phrase-level evaluation of the 7-classes subtask. This system uses morphological features of Farsi words together with the BERT model and Bi-LSTM.

\bibliography{NSURL-task7}
\bibliographystyle{acl_natbib}

\end{document}